\documentclass[11pt]{article}
\usepackage{graphicx}
\usepackage{enumerate}
\usepackage{setspace}
\usepackage{boxedminipage}
\usepackage{fancybox}
\usepackage{wrapfig}
\usepackage{alltt}
\usepackage{here}
\usepackage{multirow}
\usepackage{url}
\usepackage{mathptmx}
\usepackage{verbatim}
\usepackage{caption}
\title{A Suffix Tree Approach To Email Filtering}
\author{Rajesh M. Pampapathi, Boris Mirkin, Mark Levene \\
            rajesh@dcs.bbk.ac.uk, mirkin@dcs.bbk.ac.uk, mark@dcs.bbk.ac.uk}
\begin{document}
\maketitle \abstract
\begin{center}
We present an approach to email filtering based on the suffix tree
data structure. A method for the scoring of emails using the
suffix tree is developed and a number of scoring and score
normalisation functions are tested. Our results show that the
character level representation of emails and classes facilitated
by the suffix tree can significantly improve classification
accuracy when compared with the currently popular methods, such as
naive Bayes. We believe the method can be extended to the
classification of documents in other domains.
\end{center}

\bibliographystyle{plain}

\section{Introduction}
\label{Section Introduction}

Just as email traffic has increased over the years since its
inception, so has the proportion that is unsolicited; some
estimations have placed the proportion as high as 60\%, and the
average cost of this to business at around \$2000 per year, per
employee (see \cite{UNSPAMSTATS04} for a range of numbers and
statistics on spam). Unsolicited emails -- commonly know as
\textit{spam} -- have thereby become a daily feature of every
email user's inbox; and regardless of advances in email filtering,
spam continues to be a problem in a similar way to computer
viruses which constantly reemerge in new guises. This leaves the
research community with the task of continually investigating new
approaches to sorting the welcome emails (known as \textit{ham})
from the unwelcome spam.

We present just such an approach to email classification and
filtering based on a well studied data structure, the suffix tree
(see \cite{lloyd00monash} for a brief introduction). The approach
is similar to many existing ones, in that it uses training
examples to construct a model or profile of the class and its
features, then uses this to make decisions as to the class of new
examples; but it differs in the depth and extent of the anaysis.
For a good overview of a number of text classification methods, see
\cite{sebastiani02machine,aas99text,wiess05textmining}.

Using a suffix tree, we are able to compare not only single words,
as in most current approaches, but substrings of an arbitrary
length. Comparisons of substrings (at the level of characters) has
particular benefits in the domain of spam classification because
of the methods spammers use to evade filters. For example, they
may disguise the nature of their messages by interpolating them
with meaningless characters, thereby fooling filters based on
keyword features into considering the words, sprinkled with random
characters, as completely new and unencountered. If we instead
treat the words as character strings, and not features in
themselves, we are still able to recognise the substrings, even if
the words are broken.

Section~\ref{Section Types of Spam} gives examples of some of the
methods spammers use to evade detection which make it useful to
consider character level features. Section~\ref{Section Naive
Bayesian Classification} gives a brief explanation of the naive
Bayes method of text classification as an example of a
conventional approach. Section~\ref{Section Suffix Tree
Classification} briefly introduces suffix trees, with some
definitions and notations which are useful in the rest of the
paper, before going on to explain how the suffix tree is used to
classify text and filter spam. Section~\ref{Section Experimental
Setup} describes our experiments, the test parameters and details
of the data sets we used. Section~\ref{Section Results} presents
the results of the experiments and provides a comparison with
results in the literature. Section~\ref{Section Conclusion}
concludes.

\section{Examples of Spam}
\label{Section Types of Spam}

Spam messages typically advertise a variety of products or
services ranging from prescription drugs or cosmetic surgery to
sun glasses or holidays. But regardless of what is being
advertised, one can distinguish between the methods used by the
spammer to evade detection. These methods have evolved with the
filters which attempt to extirpate them, so there is a
generational aspect to them, with later generations becoming
gradually more common and earlier ones fading out; as this
happens, earlier generations of filters become less effective.

We present four examples of spam messages, the first of which
illustrates undisguised spam while the other three illustrate one
or more methods of evasion.
%%%%%%%%%%%%%%%%%%%%%%%%%%%%%%%%%%%%%%%%%%%%%%%%%%%%%%%%%%%%%%%%%%%%%%%%%%%
\begin{enumerate}
\item \textbf{Undisguised message.} The example contains no
obfuscation. The content of the message is easily identified by
filters, and words like ``Viagra" allow it to be recognised as
spam. Such messages are very likely to be caught by the simplest
word-based Bayesian classifiers.
\begin{flushleft}
\setlength{\fboxsep}{8pt}
\begin{boxedminipage}[0]{10cm}

Buy cheap medications online, no prescription needed. \\
We have Viagra, Pherentermine, Levitra, Soma, Ambien, Tramadol and many more products. \\
No embarrasing trips to the doctor, get it delivered directly to your door.\\

Experienced reliable service.\\

Most trusted name brands.\\

Your solution is here: http://www.webrx-doctor.com/?rid=1000 \\

\end{boxedminipage}
\end{flushleft}
%%%%%%%%%%%%%%%%%%%%%%%%%%%%%%%%%%%%%%%%%%%%%%%%%%%%%%%%%%%%%%%%%%%%%%%%%%%
\item \textbf{Intra-word characters.}
\begin{flushleft}
\setlength{\fboxsep}{12pt}
\begin{boxedminipage}[0]{10cm}

Get the  low.est  pri.ce  for  gen.eric  medica.tions!

Xa.n.ax  - only \$100 \\
Vi.cod.in  - only \$99 \\
Ci.al.is  - only \$2 per  do.se \\
Le.vit.ra  - only \$73 \\
Li.pit.or  - only \$99 \\
Pr.opec.ia  - only \$79 \\
Vi.agr.a  - only \$4 per  do.se \\
Zo.co.r  - only \$99 \\

Your  Sav.ings 40\% compared Average Internet Pr.ice! \\

No  Consult.ation  Fe.es!  No  Pr.ior  Prescrip.tions   Required!  No  Appoi.ntments! \\
No  Wait.ing  Room!  No  Embarra.ssment!  Private and  Confid.ential!  Disc.reet  Packa.ging! \\

che ck   no w:\\
http://priorlearndiplomas.com/r3/?d=getanon \\

\end{boxedminipage}
\end{flushleft}
The example above shows the use of intra-word characters, which
may be non-alpha-numeric or whitespace. Here the word, ``Viagra"
has become ``Vi.agr.a", while the word ``medications" has become
``medica.tions". To a simple word-based Bayesian classifier, these
are completely new words, which might have occurred rarely, or not
at all, in previous examples. Obviously, there are a large number
of variations on this theme which would each time create an
effectively new word which would not be recognised as spam
content. However, if we approach this email at the character
level, we can still recognise strings such as ``medica" as
indicative of spam, regardless of the character that follows, and
furthermore, though we do not deal with this in the current paper,
we might implement a look-ahead window which attempts to skip (for
example) non-alphabetic characters when searching for spammy
features.

Certainly, one way of countering such techniques of evasion is to map the
obfuscated words to genuine words during a pre-processing stage,
and doing this will help not only word-level filters, but also
character-level filters because an entire word match, either as a
single unit, or a string of characters, is better than a partial
word match.

However, some other methods may not be evaded so easily in the same way, with
each requiring its own special treatment; we give two more examples below which illustrate the point.

\item \textbf{Word salad.}
\begin{flushleft}
\setlength{\fboxsep}{12pt}
\begin{boxedminipage}[0]{10cm}

Buy meds online and get it shipped to your door Find out more here\\
http://www.gowebrx.com/?rid=1001 \\

a publications website accepted definition. known are can Commons the be definition. Commons UK great public principal work Pre-Budget but an can Majesty's many contains statements statements titles (eg includes have website. health, these Committee Select undertaken described may publications \\

\end{boxedminipage}
\end{flushleft}
The example shows the use of what is sometimes called a
\textit{word salad} - meaning a random selection of words. The
first two lines of the message are its real content; the paragraph
below is a paragraph of words taken randomly from what might have
been a government budget report. The idea is that these words are
likely to occur in ham, and would lead a traditional algorithm
to classify this email as such. Again, approaching this at the
character level can help. For example, say we consider strings of length 8,
strings such as ``are can" and ``an can", are unlikely to occur in
ham, but the words ``an", ``are" and ``can" may occur quite
frequently. Of course, in most 'bag-of-words' implementations,
words such as these are pruned from the feature set, but the
argument still holds for other bigrams. \item \textbf{Embedded
message (also contains a word/letter salad).} The example below
shows an \textit{embedded} message. Inspection of it will reveal
that it is actually offering prescription drugs. However, there
are no easily recognised words, except those that form the word
salad, this time taken from what appear to be dictionary entries
under 'z'. The value of substring searching is highly apparent in
this case as it allows us to recognise words such as ``approved",
``Viagra" and ``Tablets", which would otherwise be lost among the
characters pressed up against them.
\begin{flushleft}
\setlength{\fboxsep}{12pt}
\begin{boxedminipage}[0]{10cm}

zygotes zoogenous zoometric zygosphene zygotactic zygoid zucchettos zymolysis zoopathy zygophyllaceous zoophytologist zygomaticoauricular zoogeologist zymoid zoophytish zoospores zygomaticotemporal zoogonous zygotenes zoogony zymosis zuza zoomorphs zythum zoonitic zyzzyva zoophobes zygotactic zoogenous zombies zoogrpahy zoneless zoonic zoom zoosporic zoolatrous zoophilous zymotically zymosterol \\

FreeHYSHKRODMonthQGYIHOCSupply.IHJBUMDSTIPLIBJT \\

\verb"* GetJIIXOLDViagraPWXJXFDUUTabletsNXZXVRCBX" \\
http://healthygrow.biz/index.php?id=2 \\

zonally zooidal zoospermia zoning zoonosology zooplankton zoochemical zoogloeal zoological zoologist zooid zoosphere zoochemical \\

\verb"& Safezoonal andNGASXHBPnatural" \\
\verb"& TestedQLOLNYQandEAVMGFCapproved" \\

zonelike zoophytes zoroastrians zonular zoogloeic zoris zygophore zoograft zoophiles zonulas zygotic zymograms zygotene zootomical zymes zoodendrium zygomata zoometries zoographist zygophoric zoosporangium zygotes zumatic zygomaticus zorillas zoocurrent zooxanthella zyzzyvas zoophobia zygodactylism zygotenes zoopathological noZFYFEPBmas http://healthygrow.biz/remove.php \\

\end{boxedminipage}
\end{flushleft}

\end{enumerate}

\vspace{0.5cm}

These examples are only a sample of all the types of spam that
exist, for an excellent and often updated list of examples and
categories, see \cite{jgc04compendium,levene04spam}. Under the
categories suggested in \cite{wittel04attack}, example 2 and 4
would count as 'Tokenisation' and/or 'Obfuscation', while examples
2 and 3 would count as 'Statistical'.

We look next at a bag-of-words approach, naive Bayes, before
considering the suffix tree approach.

\section{Naive Bayesian Classification}
\label{Section Naive Bayesian Classification}

Naive Bayesian (NB) email filters currently attract a lot of
research and commercial interest, and have proved highly
successful at the task; \cite{sahami98bayesian} and
\cite{androutsopoulos04filtron} are both excellent studies of this
approach to email filtering. We do not give detailed attention to
NB as it is not the intended focus of this paper; for a general
discussion of NB see~\cite{lewis98naive}, for more context in text
categorisation see~\cite{sebastiani02machine}, and for an
extension of NB to the classification of structured data,
see~\cite{Flach05NBStructured}. However, an NB classifier is
useful in our investigation of the suffix tree classifier, and in
particular, our own implementation of NB is necessary to
investigate experimental conditions which have not been explored in
the literature. We therefore briefly present it here.

We begin with a set of training examples with each example
document assigned to one of a fixed set of possible classes,
\textbf{C} = \{c$_{1}$, c$_{2}$, c$_{3}$,... c$_{J}$\}. An NB
classifier uses this training data to generate a probabilistic
model of each class; and then, given a new document to classify,
it uses the class models and Bayes' rule to estimate the
likelihood with which each class generated the new document. The
document is then assigned to the most likely class. The features,
or parameters, of the model are individual words; and it is
'naive' because of the simplifying assumption that, given a class,
each parameter is independent of the others.

\cite{mccallum98comparison} distinguish between two types of probabilistic models which are commonly used in NB classifiers: the \textit{multi-variate Bernoulli} event model and the \textit{multinomial} event model. We adopt the latter, under which a document is seen as a series of word events and the probability of the event given a class is estimated from the frequency of that word in the training data of the class.

Hence, given a document $\textbf{d}=\{d_1d_2d_3 ... d_L\}$, we use Bayes theorem to
estimate the probability of a class, \textit{c$_{j}$}:
\begin{equation}
P(c_{j} \mid \mathbf{d}) = \frac{P(c_{j})P(\mathbf{d} \mid
c_{j})}{P(\mathbf{d})}
\end{equation}
Assuming that words are independent given the category, this leads
to:
\begin{equation}
P(c_{j} \mid \mathbf{d}) = \frac{P(c_{j})\prod_{i=1}^{L}P(d_{i}
\mid c_{j})}{P(\mathbf{d})}
\end{equation}
We estimate P(c$_{j}$) as:
\begin{equation}
\hat{P}(C = c_{j}) = \frac{\mathit{N_{j}}}{\mathit{N}}
\end{equation}
and P(d$_{i}$ $\mid$ c$_{j}$) as:
\begin{equation}
\hat{P}(d_{i}\mid c_{j}) = \frac{1 + \mathit{N_{ij}}}{\mathit{M} +
\sum_{k=1}^{M}\mathit{N_{kj}}} \label{P(d_i|c_j)}
\end{equation}
where $N_{ij}$ is the number of times word $i$ occurs in class $j$
(similarly for $N_{kj}$) and $M$ is the total number of words
considered.

To classify a document we calculate two scores, for spam and ham,
and take the ratio, $hsr = \frac{hamScore}{spamScore}$, and
classify the document as ham if it is above a threshold, $th$, and
as spam if it is below (see Section~\ref{SubSubSection
Threshold}).

\section{Suffix Tree Classification}
\label{Section Suffix Tree Classification}

\subsection{Introduction}
\label{SubSection Introduction}

The suffix tree is a data storage and fast search technique which
has been used in fields such as computational biology for
applications such as string matching applied to DNA
sequences~\cite{Bejerano_Gill2001,Bingwen2003}. To our
knowledge it has not been used in the domain of natural language
text classification.

We adopted a conventional procedure for using a suffix tree in
text classification. As with NB, we take a set of documents
$\bf{D}$ which are each known to belong to one class, $c_j$, in a
set of classes, $\bf{C}$, and build one tree for each class. Each
of these trees is then said to represent (or profile) a class (a
tree built from a class will be referred to as a ``class tree").

Given a new document $\bf{d}$, we score it with respect to each of
the class trees and the class of the highest scoring tree is taken
as the class of the document.

We address the scoring of documents in Section~\ref{SubSection
Classification using Suffix Trees}, but first, we consider the
construction of the class tree.

\subsection{Suffix Tree Construction}
\label{SubSection Suffix Tree Construction}

\setlength{\captionmargin}{40pt}
\begin{figure}
\begin{center}
\includegraphics[width = 6cm]{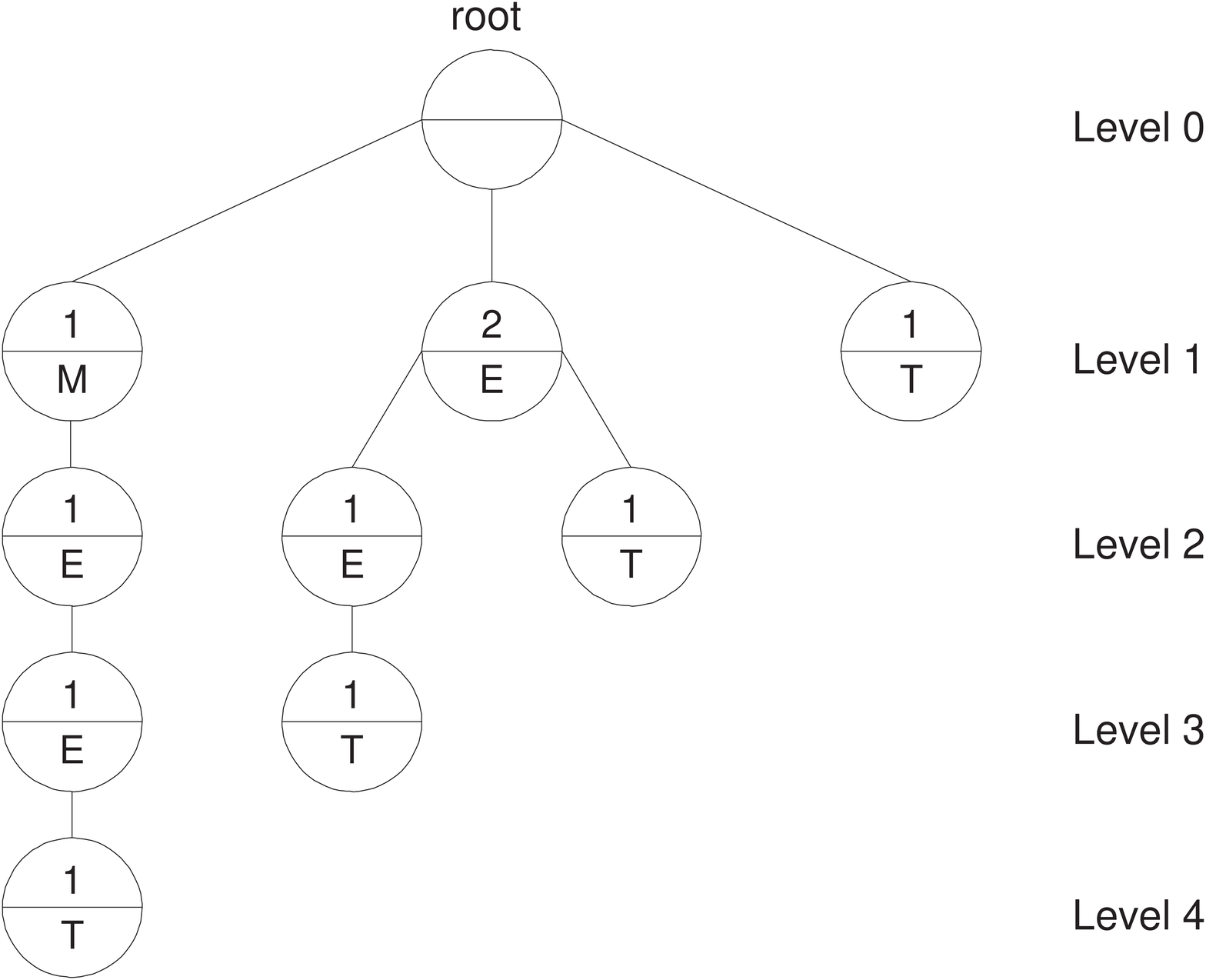}
\caption{A Suffix Tree after the insertion of ``meet".}
\label{Figure_STMEET}
\end{center}
\end{figure}
\setlength{\captionmargin}{20pt}
\begin{figure}
\begin{center}
\includegraphics[width = 7.37cm]{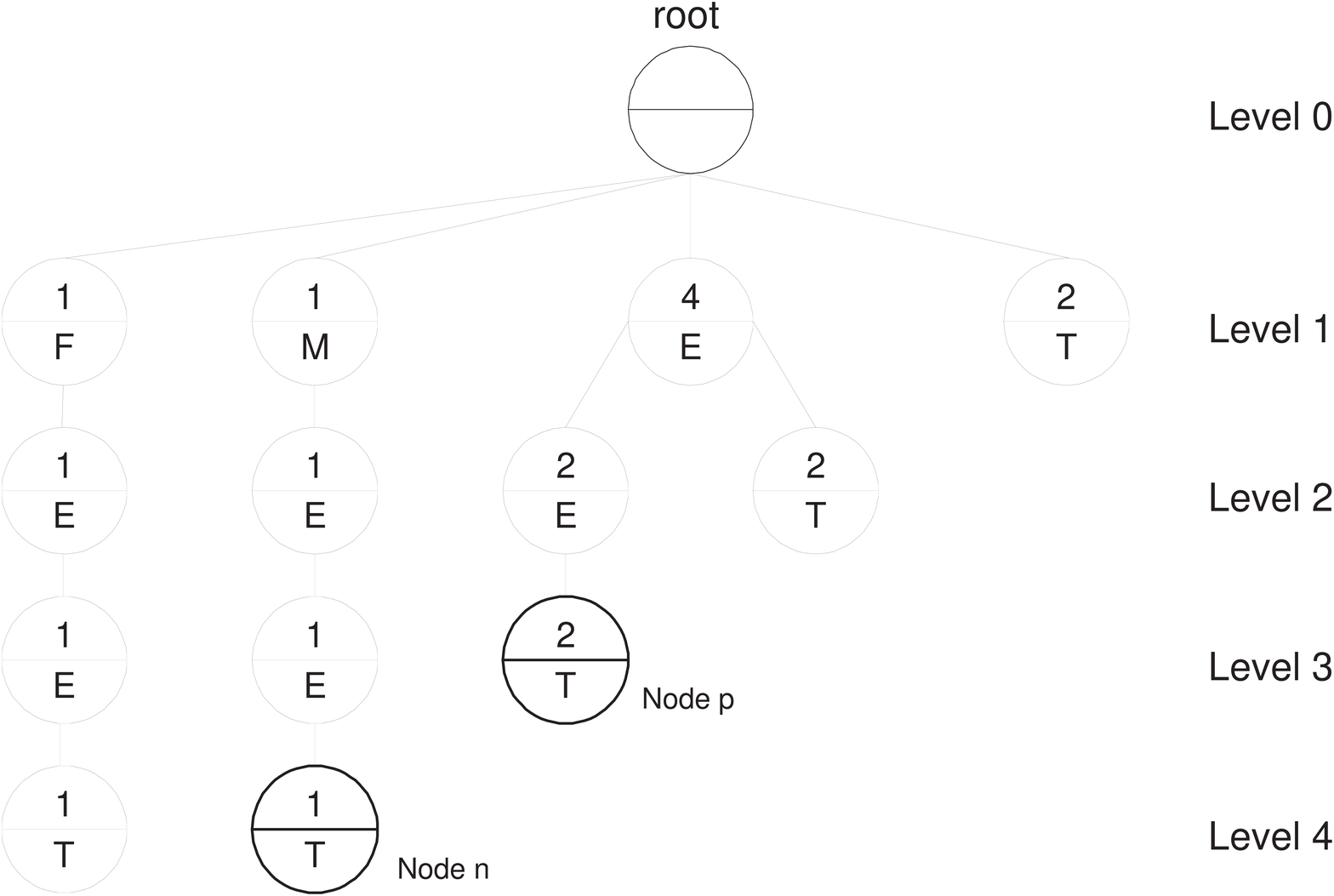}
\caption{A Suffix Tree after insertion of strings ``meet" and
``feet".} \label{Figure_STMEET_FEET}
\end{center}
\end{figure}

We provide a brief introduction to suffix tree construction. For a
more detailed treatment, along with algorithms to improve
computational efficiency, the reader is directed to
\cite{gusfield97algorithms}. Our representation of a suffix tree
differs from the literature in two ways that are specific to our
task: first, we label nodes and not edges, and second, we do not
use a special terminal character. The former has little impact on
the theory and allows us to associate frequencies directly with
characters and substrings. The later is simply because our
interest is actually focused on substrings rather than suffixes;
the inclusion of a terminal character would therefore not aid our
algorithms, and its absence does not hinder them. Furthermore, our
trees are depth limited, and so the inclusion of a terminal
character would be meaningless in most situations.

Suppose we want to construct a suffix tree from the string, $s =
``meet"$. The string has four suffixes: $s(1) = ``meet"$, $s(2) =
``eet"$, $s(3) = ``et"$, and $s(4) = ``t"$.

We begin at the root of the tree and create a child node for the
first character of the suffix $s(1)$. We then move down the tree
to the newly created node and create a new child for the next
character in the suffix, repeating this process for each of the
characters in this suffix. We then take the next suffix, $s(2)$,
and, starting at the root, repeat the process as with the previous
suffix. At any node, we only create a new child node if none of
the existing children represents the character we are concerned
with at that point. When we have entered each of the suffixes, the
resulting tree looks like that in Figure 1. Each node is labelled
with the character it represents and its frequency. The node's
position also represents the position of the character in the
suffix, such that we can have several nodes labelled with the same
character, but each child of each node (including the root) will
carry a character label which is unique among its siblings.

If we then enter the string, $t = ``feet"$, into the tree in
Figure~\ref{Figure_STMEET}, we obtain the tree in
Figure~\ref{Figure_STMEET_FEET}. The new tree is almost identical
in structure to the previous one because the suffixes of the two
strings are all the same but for $t(1) = ``feet"$, and as we said
before, we need only create a new node when an appropriate node
does not already exist, otherwise, we need only increment the
frequency count.

Thus, as we continue to add more strings to the tree, the number
of nodes in the tree increases only if the new string contains
substrings which have not previously been encountered. It follows
that given a fixed alphabet and a limit to the length of
substrings we consider, there is a limit to the size of the tree.
Practically, we would expect that, for most classes, as we
continue to add strings to the class tree, the tree will increase
in size at a decreasing rate, and will quite likely stabilise.

\subsection{Class Trees and their Characteristics}
\label{SubSection Class Trees and their Characteristics}

For any string $s$ we designate the $i^{th}$ character of $s$ by
$s_i$; the suffix of $s$ beginning at the $i^{th}$ character by
$s(i)$; and the substring from the $i^{th}$ to the $j^{th}$
character inclusively by $s(i,j)$.

Any node, $n$, labelled with a character, $c$, is uniquely
identified by the path from the $root$ to $n$. For example,
consider the tree in Figure~\ref{Figure_STMEET_FEET}. There are
several nodes labelled with a ``$t$", but we can distinguish
between node $n = (``t" \hspace{1ex} given \hspace{1ex} ``mee") =
(t|mee)$ and $p = (``t" \hspace{1ex} given \hspace{1ex} ``ee") =
(t|ee)$; these nodes are labelled $n$ and $p$ in
Figure~\ref{Figure_STMEET_FEET}. We say that the path of $n$ is
$\overrightarrow{P}_n = ``mee"$, and the path of $p$ is
$\overrightarrow{P}_p = ``ee"$; furthermore, the frequency of $n$
is 1, whereas the frequency of $p$ is 2; and saying $n$ has a
frequency of 1, is equivalent to saying the frequency of ``$t$"
given ``$mee$" is 1, and similarly for $p$.

If we say that the $root$ node, $r$, is at level zero in the tree,
then all the children of $r$ are at level one. More generally, we
can say that the level of any node in the tree is one plus the
number of letters in its path. For example, $level(n)$ = 4 and
$level(p)$ = 3.

The set of letters forming the first level of a tree is the
\textit{alphabet}, $\Sigma$ - meaning that all the nodes of the
tree are labelled with one of these letters. For example,
considering again the tree in Figure~\ref{Figure_STMEET_FEET}, its
first level letters are the set, $\Sigma = \{m, e, t, f\}$, and
all the nodes of the tree are labelled by one of these.

Suppose we consider a class, $C$, containing two strings (which we
might consider as documents), $s = ``meet"$ and $t = ``feet"$.
Then we can refer to the tree in Figure~\ref{Figure_STMEET_FEET}
as the \textit{class tree} of $C$, or the \textit{suffix tree
profile} of $C$; which we denote by $T_C$.

The size of the tree, $|T_C|$, is the number of nodes it has, and
it has as many nodes as $C$ has unique substrings. For instance,
in the case of the tree in Figure~\ref{Figure_STMEET_FEET}:
%\vspace{0.1cm}

\[ UC = \texttt{uniqueSubstrings}(C) = \left\{
                                              \begin{array}{ll}
                                               meet, mee, me, m, eet, ee, e, et, t, \\
                                               feet, fee, fe, f
                                              \end{array}
                                       \right\}
\]

\vspace{0.05cm}
$|UC| = |\texttt{uniqueSubstrings}(C)| = 13$ \\

\vspace{0.05cm}
$|T_C| = \texttt{numberOfNodes}(T_C) = 13$ \\
%\vspace{0.2cm}

This is clearly not the same as the total number of substrings
(tokens) in $C$:

\[ AC = \texttt{allSubstrings}(C) =  \left\{
                                            \begin{array}{ll}
                                            meet, mee, me, m, eet, ee, e, et, e, t, \\
                                            feet, fee, fe, f, eet, ee, e, et, e, t
                                            \end{array}
                                     \right\}
\]

\vspace{0.05cm}
$|AC| = |\texttt{AllSubstrings}(C)| = 20$ \\

\vspace{0.2cm}

As an example, note that the four ``e"s in the set are in fact the
substrings s(1,1), s(2,2), t(1,1) and t(2,2).

Furthermore, as each node in the tree, $T_C$, represents one of
the substrings in UC, the size of the class, AC, is equal to the
sum of the frequencies of nodes in the tree $T_C$. \\

$|AC| = |\texttt{allSubstrings}(C)| =
\texttt{sumOfFrequencies}(T_C) = 20$ \\ \vspace{0.2cm}

In a similar way, the suffix tree allows us to read off other
frequencies very quickly and easily. For example, if we want to
know the number of characters in the class $C$, we can sum the
frequencies of the nodes on the first level of the tree; and if we
want to know the number of substrings of length 2, we can sum the
frequencies of the level two nodes; and so on.

This also allows us to very easily estimate probabilities of
substrings of any length (up to the depth of the tree), or of any
nodes in the tree. For example, we can say from the tree in
Figure~\ref{Figure_STMEET_FEET}, that the probability of a
substring, $u$, of length two, having the value, $u = ``ee"$,
given the class $C$, is the frequency, $f$, of the node $n =
(e|e)$, divided by the sum of the frequencies of
all the level two nodes in the tree $T_C$:\\
\vspace{0.1cm}
\begin{equation}
\texttt{estimatedTotalProbability}(u) = {\Large\frac{f(u)}{\sum_{i \in N_u}f(i)}} \\
\label{equation: estimatedProbabilityOfString}
\end{equation}
\vspace{4ex}where $N_u$ is the set of all nodes at same level as $u$.\\
\vspace{0.1cm}Similarly one can estimate the conditional
probability of $u$ as the frequency of $u$ divided by the sum of
the
frequencies of all the children of $u$'s parent:\\
\vspace{0.1cm}
\begin{equation}
\texttt{estimatedConditionalProbability}(u) = {\Large\frac{f(u)}{\sum_{i \in n_u}f(i)}} \\
\label{equation: estimatedProbabilityOfChar}
\end{equation}
\vspace{0.2cm}where $n_u$ is the set of all children of $u$'s parent.\\

\vspace{0.1cm} Throughout this paper, whenever we mention
$\hat{p}(u)$, we mean the second of these (formula~(\ref{equation:
estimatedProbabilityOfChar})): the \textit{conditional
probability} of a node $u$.

\subsection{Classification using Suffix Trees}
\label{SubSection Classification using Suffix Trees}

Researchers have tackled the problem of the construction of a text
classifier in a variety of different ways, but it is popular to
approach the problem as one that consists of two parts:
\begin{enumerate}
\item \label{Item Scoring Function} The definition of a function,
$CS_i:D\rightarrow\mathbf{R}$, where $D$ is the set of all
documents; such that, given a particular document, $d$, the
function returns a \textit{category score} for the class $i$. The
score is often normalised to ensure that it falls in the the
region $[0,1]$, but this is not strictly necessary, particularly
if one intends, as we do, simply to take as the class prediction
the highest scoring class (see Part~\ref{Item ST Decision
Mechanism} below). The interpretation of the meaning of the
function, $CS$, depends on the approach adopted. For example, as
we have seen, in naive Bayes, $CS(d)$, is interpreted as a
\textit{probability}; whereas in other approaches such as
Rocchio~\cite{Rocchio71}, $CS(d)$ is interpreted as a
\textit{distance} or \textit{similarity} measure between two
vectors.

\item \label{Item ST Decision Mechanism}A decision mechanism which
determines a class prediction from set of class scores. For
example, the highest scoring class might be taken as the predicted
class: $PC = argmax_{c_j \in C}\{CS_j(d)\}$. Alternatively, if
$CS(d)$ is interpreted as a value with definite range, such as a
probability, the decision may be based on a threshold, $th$, such
that the predicted class is taken as $c_j$ if $CS_j(d) > th$, and
as not $c_j$ otherwise.
\end{enumerate}

\cite{lewis96training,duda73classification} refer to probabilistic models such as naive
Bayes as parametric classifiers because they attempt to use the
training data to estimate the parameters of a probability
distribution, and assume that the estimated distribution is
correct. Non-parametric, geomtric models, such as Rocchio~\cite{Rocchio71},
instead attempt to produce a profile or summary of the training
data and use this profile to query new documents to decide their
class.

It is possible to approach the construction of a suffix tree
classifier in either of these two ways and indeed a
probability-based approach has been developed by
\cite{Bejerano_Gill2001} for use in gene sequence matching.
However, \cite{Bejerano_Gill2001} did not find the suffix tree
entirely convenient for developing a probabilistic framework and
instead developed a probabilistic analogue to the suffix tree and
used this modified data structure to develop probabilistic
matching algorithms.

In this paper, we retain the original structure of the suffix tree
and favour a non-parametric, or geometric, approach to classifier
construction. In such a framework a match between a document and a
suffix tree profile of a class is a set of coinciding substrings
each of which must be scored individually so that the total score
is the sum of individual scores. This is analogous to the inner
product between a document vector and class profile vector in the
Rocchio algorithm~\cite{Rocchio71}. We did experiment with probabilistic models,
and found that it was possible to construct one without altering
the structure of the suffix tree (indeed, some of the flavours of
the scoring system we present can be seen as approximating a
probabilistic approach (see Section~\ref{SubSubSection Scoring},
Part~\ref{Item Significance function}) even though the branches
are not independent: each corresponds to a set of strings which
may overlap. However we found that additive scoring algorithms
performed better and in the current paper we describe only this
approach. The method, the details of which are presented in the
next section, is governed by two heuristics:
\begin{enumerate}
\item[H1] \label{heuristic: substring match length} Each substring
$s(i)$ that a string $s$ has in common with a class $T$ indicates
a degree of similarity between $s$ and $T$, and the longer the
common substrings the greater the similarity they indicate.
\item[H2] \label{heuristic: class diversity}The more
diverse\footnote{\textit{Diversity} is here an intuitive notion
which the scoring method attempts to define and represent in a
number of different ways.} a class $T$, the less significant is the
existence of a particular common substring $s(i,j)$ between a string
$s$ and the class $T$ .
\end{enumerate}

Turning to the second issue in classifier construction, for our
current two-class problem, we take the ratio of two scores, $hsr =
\frac{hamScore}{spamScore}$, just as we did in the case of our
naive Bayesian classifier, and classify the document as ham if the
ratio is greater than a threshold, $th$, and as spam if the ratio
is below $th$. By raising and lowering this threshold we can
change the relative importance we place on miss-classified spam
and ham messages (see Section~\ref{SubSubSection Threshold}).

\subsubsection{Scoring}
\label{SubSubSection Scoring}

The suffix tree representation of a class is richer than the
vector representation of more traditional approaches and in
developing a scoring method we can experiment with various
properties of the tree, each of which can be seen as reflecting
certain properties of the class and its members.

We begin by describing how to score a \textit{match} between a
string and a class, then extend this to encompass document
scoring. Conceptually dividing the scoring in this way allowed us
to introduce and experiment with two levels of normalisation:
match-level, reflecting information about strings; and tree-level,
reflecting information about the class as a whole. The end of this
section elaborates on the underlying motivation for the described
scoring method.

\begin{enumerate}
\item \textbf{Scoring a match} \label{Item Scoring a match}
\begin{enumerate}
\item We define a match as follows: A string $s$ has a match $m =
m(s, T)$ in a tree $T$ if there exists in $T$ a path
$\overrightarrow{P} = m$, where $m$ is a prefix of $s$.

\hspace{3ex}Clearly, the match $m$ may represent several substrings
that are common between $s$ and $T$. However, it is important to
note that if $|m| > 1$ and $m_0 = s_i$, then we would expect to
find another match $m'$ beginning at $s_{i+1}$, such that $|m'|
\geq |m| - 1$, hence we can think of $m$ as representing only
those substrings common to both $s$ and $T$ which begin with $m_0$
and still be sure that the set of all matches, $M$, between $s$
and $T$ will represent each common substring.

\item The score, $score(m)$, for a match $m = m_0m_1m_2 ... m_n$,
has two parts, firstly, the scoring of each character (and
thereby, each substring), $m_i$, with respect to its conditional
probability, using a \textit{significance} function of
probability, $\phi[\hat{p}]$ (defined below in part(\ref{Item
Significance function})), and secondly, the adjustment
(normalisation), $v(m|T)$, of the score for the whole match with
respect to its probability in the tree:
\begin{equation}
score(m) = \nu(m|T)\sum_{i=0}^n \phi[\hat{p}(m_i)]
\label{equation: score(m)}
\end{equation}
Using the conditional probability  rather than the total
probability has the benefit of supporting heuristic H1: as we go
deeper down the tree, each node will tend to have fewer children
and so the conditional probability will be likely to increase;
conversely, there will generally be an increasing number of nodes at
each level and so the total probability of a particular node will
decrease. Indeed we did experiment with the total probability and
found that performance was significantly decreased.

\hspace{3ex}Furthermore, by using the conditional probability we
also only consider the independent parts of features when deriving
scores. So for example, if $m = ``abc"$, by the time we are
scoring the feature represented by ``\textit{abc}", we have
already scored the feature ``\textit{ab}", so we need only score
``\textit{c}" given ``\textit{ab}".

\item \label{Item Significance function} A function of
probability, $\phi[\hat{p}]$, is employed as a
\textit{significance function} because it is not always the most
frequently occurring terms or strings which are most indicative of
a class. For example, this is the reason that conventional
pre-processing removes all stop words, and the most and least
frequently occurring terms; however, by removing them completely
we give them no significance at all, when we might instead include
them, but reduce their significance in the classification
decision. Functions on the probability can help to do this,
especially in the absence of all pre-processing, but that still
leaves the question of \textit{how} to weight the probabilities,
the answer to which will depend on the class.

\hspace{3ex}In the spam domain, some strings will occur very
infrequently (consider some of the strings resulting from
intra-word characters in the examples of spam in Section~\ref{Section Types of Spam} above) in either the
spam or ham classes, and it is because they are so infrequent that
they are indicative of spam. Therefore, under such an argument,
rather than remove such terms or strings, we should actually
\textit{increase} their weighting.

\hspace{3ex}Considerations such as these led to experimentation
with a number of specifications of the significance function,
$\phi[\hat{p}]$:
               \[   \phi [\hat{p}] = \left\{
                                   \begin{array}{ll}
                                         1 & \hspace{1pt} constant \vspace{1ex} \\
                                         \hat{p} & \hspace{1pt} linear \vspace{1ex} \\
                                         \hat{p}^{\hspace{1pt}2} & \hspace{1pt} square \vspace{1ex} \\
                                         \sqrt{\hat{p}} & \hspace{1pt} root \vspace{1ex} \\
                                         \ln(\hat{p}) - \ln(1 - \hat{p}) & \hspace{1pt} logit \vspace{1ex} \\ %\frac{1}{B}\{\ln(T) - \ln[(\frac{C}{\hat{p} + A})^T - 1]\} + M
                                         \frac{1}{1 + \exp(-\hat{p})} & \hspace{1pt} sigmoid \vspace{1ex} \\ %A + C\{1 + T\exp(-B(\hat{p} - M))\}^{-\frac{1}{T}}
                                   \end{array}
                                   \right.
               \]
The first three functions after the constant are variations of the
linear (linear, sub-linear and super-linear). The last two are
variations on the S-curve; we give above the simplest forms of the
functions, but in fact, they must be adjusted to fit in the range
[0,1].

\hspace{3ex}Although in this paper we are not aiming to develop a
probabilistic scoring method, note that the logistic significance
function applied to formula (\ref{equation: score(m)}) may be
considered an approximation of such an approach since we generally have a
large alphabet and therefore a large number of children at each
node, and so for most practical purposes $\ln(1-\hat{p}) \approx
0$.

\item \label{Item Match normalisation}

Turning our attention to match-level normalisation, we
experimented with three specifications of $\nu(m|T)$:
\begin{center}\[ \nu(m|T) =  \left\{
                                \begin{array}{ll}
                                1 & match \hspace{1ex} \hspace{1pt} unnormalised \vspace{1ex} \\
                                \Large{\frac{f(m|T)}{\sum_{i \in (m^*|T)} f(i)}} & \hspace{1pt} match \hspace{1ex} permutation \hspace{1ex} normalised  \vspace{1ex} \\
                                \Large{\frac{f(m|T)}{\sum_{i \in (m'|T)} f(i)}} & \hspace{1pt} match \hspace{1ex} length \hspace{1ex} normalised\\
                                \end{array}
                            \right.
               \]

\end{center}
where $m^*$ is the set of all the strings in $T$ formed by the
permutations of the letters in $m$; and $m'$ is the set of all
strings in $T$ of length equal to the length of $m$.
\end{enumerate}
Match permutation normalisation (MPN) is motivated by heuristic
H2. The more diverse a class (meaning that it is represented by a
relatively large set of substring features), the more combinations
of characters we would expect to find, and so finding the
particular match $m$ is less significant than if the class were
very narrow (meaning that it is fully represented by a relatively
small set of substring features). Reflecting this, the MPN
parameter will tend towards 1 if the class is less diverse and
towards 0 if the class is more diverse.

\hspace{3ex}Match length normalisation (MLN) is motivated by
examples from standard linear classifiers (see
\cite{lewis96training} for an overview), where length
normalisation of feature weights is not uncommon. However, MLN
actually runs counter to heuristic H1 because it will tend towards
0 as the match length increases. We would therefore expect MLN to
reduce the performance of the classifier; thus MLN may serve as a
test of the intuitions governing heuristic H1.

\item \textbf{Scoring a document} \label{Item Scoring a Document
Main}
\begin{enumerate}
\item \label{Item Scoring a Document Sub} To score an entire
document we consider each suffix of the document in turn and score
any match between that suffix and the class tree. Thus the score
for a document $s$ is the sum:
\begin{equation}
SCORE(s, T) = \hspace{1ex}\sum_{i=0}^n score(s(i), T)
\label{equation: SCORE(d, T)}
\end{equation}

where the $score(s(i), T)$ searches for a match, $m$, between
suffix $s(i)$ and tree $T$, and if one is found, scores it
according to formula~(\ref{equation: score(m)}).

\hspace{3ex}We experimented with a number of approaches to
tree-level normalisation of the sum in (\ref{equation: SCORE(d,
T)}) motivated again by heuristic H2 and based on tree properties
such as $size$, as a direct reflection of the diversity of the
class; $density$ (defined as the average number of children over
all internal nodes), as an implicit reflection of the diversity;
and total and average $frequencies$ of nodes as an indication of
the size of the class\footnote{Class size is defined as the total
number of substrings in the documents of the class, and tree size
as the number of nodes in the tree, that is, the number of unique
substrings in the class (see Section~\ref{SubSection Class Trees
and their Characteristics}).}; but found none of our attempts to
be generally helpful to the performance of the classifier.
Unfortunately, we do not have space in this paper to further
discuss this aspect.

\end{enumerate}
\end{enumerate}

The underlying mechanism of the scoring function can be grasped by
considering its simplest configuration: using the constant
significance function, with no normalisation. If the scoring
method were used in this form to score the similarity between two
strings, it would simply count the number of substrings that the
two strings have in common. For example, suppose we have a string
$t = ``abcd"$. If we were to apply this scoring function to
assessing the similarity that $t$ has with itself, we would obtain
a result of 11, because this is the number of unique substrings
that exist in $t$. If we then score the similarity between $t$ and
$t^0 = ``Xbcd"$, we obtain a score of 6, because the two strings
share 6 unique substrings; similarly, a string $t^1 = ``aXcd"$
would score 4.

Another way of viewing this is to think of each substring of $t$
as representing a feature in the class that $t$ represents. The scoring method then weights each of these as 1 if they are present
in a query string and 0 otherwise, in a way that is analogous to
the simplest form of weighting in algorithms such as Rocchio.

Once seen in this way, we can consider all other flavours of the
classifier as experimenting with different approaches to deciding
how significant each common substring is, or in other words,
deciding how to weight each class feature -- in much the same as
with other \textit{non-parametric} classifier algorithms.

\section{Experimental Setup}
\label{Section Experimental Setup}

All experiments were conducted under ten-fold cross validation. We
accept the point made by~\cite{meyer04spambayes} that such a
method does not reflect the way classifiers are used in practice,
but the method is widely used and serves as a thorough initial
test of new approaches.

We follow convention by considering as true positives (TP),
spam messages which are correctly classified as spam; false
positives (FP) are then ham messages which are incorrectly
classified as spam; false negatives (FN) are spam incorrectly
classified as ham; true negatives (TN) are ham messages correctly
classified as ham. See Section~\ref{SubSection Performance
Measurement} for more on the performance measurements we use.

\subsection{Experimental Parameters}
\label{SubSection Experimental Parameters}
\subsubsection{Spam to Ham Ratios} \label{SubSubSection Spam to Ham Ratios} From some initial tests we found
that success was to some extent contingent on the proportion of
spam to ham in our data set -- a point which is identified, but
not systematically investigated in other work
\cite{meyer04spambayes} -- and this therefore became part of our
investigation. The differing results further prompted us to
introduce forms of normalisation, even though we had initially
expected the probabilities to take care of differences in the
scale and mix of the data. Our experiments used three different
ratios of spam to ham: 1:1, 4:6, 1:5. The first and second of
these (1:1 and 4:6) were chosen to reflect some of the estimates
made in the literature of the actual proportions of spam in
current global email traffic. The last of these (1:5) was chosen
as the minimum proportion of spam included in experiments detailed
in the literature, for example in
\cite{androutsopoulos00evaluation}.

\subsubsection{Tree Depth} \label{SubSubSection Tree Depth} It is too computationally expensive to
build trees as deep as emails are long. Furthermore, the marginal
performance gain from increasing the depth of a tree, and
therefore the length of the substrings we consider, may be
negative. Certainly, our experiments show a diminishing marginal
improvement (see Section~\ref{SubSubSection Effect of Depth
Variation}), which would suggest a maximal performance level,
which may not have been reached by any of our trials. We
experimented with depths of length of 2, 4, 6, and 8.

\subsubsection{Threshold} \label{SubSubSection Threshold} From
initial trials, we observed that the choice of threshold value in
the classification criterion can have a significant, and even
critical, effect on performance, and so introduced it as an
important experimental parameter. We used a range of threshold
values between 0.7 and 1.3, with increments of 0.1, with a view to
probing the behaviour of the scoring system.

Varying the threshold is equivalent to associating higher costs
with either false positives or false negatives because checking that $(\alpha/\beta) > t$ is equivalent to checking that $\alpha > t\beta$.

\subsection{Data}
\label{SubSection Data} Three corpora were used to create the
training and testing sets:

\begin{enumerate}
\item \underline{The Ling-Spam corpus} (LS) \\
This is available from:
\url{http://www.aueb.gr/users/ion/data/lingspam_public.tar.gz}.
The corpus is that used in~\cite{androutsopoulos00evaluation}. The
spam messages of the corpus were collected by the authors from
emails they received. The ham messages are taken from postings on
a public online linguist bulletin board for professionals; the
list was moderated, so does not contain any spam. Such a source may
at first seem biased, but the authors claim that this is not the
case.  There are a total of 481 spam messages and 2412 ham
messages, with each message consisting of a subject and body.

When comparing our results against those of
~\cite{androutsopoulos00evaluation} in Section~\ref{SubSection Assessment} we use the complete data set,
but in further experiments, where our aim was to probe the
properties of the suffix tree approach and investigate the effect
of different proportions of spam to ham messages, we use a random
subset of the messages so that the sizes and ratios of the experimental data
sets derived from this source are the same as data sets made up of
messages from other sources (see Table~\ref{Table Composition Of
EDSs} below).
\item \underline{Spam Assassin public corpus} (SA) \\
This is available from:
\url{http://spamassassin.org/publiccorpus}. The corpus was
collected from direct donations and from public forums over two
periods in 2002 and 2003, of which we use only the later. The set
from 2003 comprise a total of 6053 messages, approximately 31\% of
which are spam. The ham messages are split into 'easy ham' (SAe)
and 'hard ham' (SAh), the former being again split into two groups
(SAe-G1 and SAe-G2); the spam is similarly split into two groups
(SAs-G1 and SAs-G2), but there is no distinction between hard and
easy. The compilers of the corpus describe hard ham as being
closer in many respects to typical spam: use of HTML, unusual HTML
markup, coloured text, ``spammish-sounding" phrases etc..

In our experiments we use ham from the hard group and the second
easy group (SAe-G2); for spam we use only examples from the second
group (SAs-G2). Of the hard ham there are only 251 emails, but for
some of our experiments we required more examples, so whenever
necessary we padded out the set with randomly selected examples
from group G2 of the easy ham (SAe-G2); see Table~\ref{Table
Composition Of EDSs}. The SA corpus reproduces all header
information in full, but for our purposes, we extracted the
subjects and bodies of each; the versions we used are available
at:
\url{http://dcs.bbk.ac.uk/~rajesh/spamcorpora/spamassassin03.zip}
\item \underline{The BBKSpam04 corpus} (BKS) \\
This is available at:
\url{http://dcs.bbk.ac.uk/~rajesh/spamcorpora/bbkspam04.zip}. This
corpus consists of the subjects and bodies of 600 spam messages
received by the authors during 2004. The Birkbeck School of
Computer Science and Information Systems uses an installation the SpamAssassin filter~\cite{spamassassin04} with default settings, so
all the spam messages in this corpus have initially evaded that filter. The
corpus is further filtered so that no two emails share more than
half their substrings with others in the corpus. Almost all the messages in this collection contain some kind of obfuscation, and so more accurately reflect the
current level of evolution in spam.
\end{enumerate}

\setlength{\captionmargin}{24pt}
\begin{table}
\begin{center}
\caption{Composition of Email Data Sets (EDSs) used in the
experiments.} \vspace{0.5cm}
\begin{tabular}{|c|c|c|}
  \hline
  EDS Code & Spam Source & Ham Source \\
  & (number from source) & (number from source) \\
  \hline
   & & \\
   LS-FULL & LS (481) & LS (2412) \\
   & & \\
  LS-11 & LS (400) & LS (400) \\
  LS-46 & LS (400) & LS (600) \\
  LS-15 & LS (200) & LS (1000) \\
   & & \\
  SAe-11 & SAs-G2 (400) & SAe-G2 (400)\\
  SAe-46 & SAs-G2 (400) & SAe-G2 (600) \\
  SAe-15 & SAs-G2 (200) & SAe-G2 (1000) \\
   & & \\
  SAeh-11 & SAs-G2 (400) & SAe-G2 (200) + SAh (200) \\
  SAeh-46 & SAs-G2 (400) & SAe-G2 (400) + SAh (200) \\
  SAeh-15 & SAs-G2 (200) & SAe-G2 (800) + SAh (200) \\
   & & \\
  BKS-LS-11 & BKS (400) & LS (400) \\
  BKS-LS-46 & BKS (400) & LS (600) \\
  BKS-LS-15 & BKS (200) & LS (1000) \\
   & & \\
  BKS-SAe-11 & BKS (400) & SAe-G2 (400)\\
  BKS-SAe-46 & BKS (400) & SAe-G2 (600) \\
  BKS-SAe-15 & BKS (200) & SAe-G2 (1000) \\
   & & \\
  BKS-SAeh-11 & BKS (400) & SAe-G2 (200) + SAh (200) \\
  BKS-SAeh-46 & BKS (400) & SAe-G2 (400) + SAh (200) \\
  BKS-SAeh-15 & BKS (200) & SAe-G2 (800) + SAh (200) \\
   & & \\
  \hline
\end{tabular}
\label{Table Composition Of EDSs}
\end{center}
\end{table}

One experimental \textit{email data set} (EDS) consisted of a set
of spam and a set of ham. Using messages from these three corpora,
we created the EDSs shown in Table~\ref{Table Composition Of
EDSs}. The final two numbers in the code for each email data set
indicate the mix of spam to ham; three mixes were used: 1:1, 4:6,
and 1:5. The letters at the start of the code indicate the source
corpus of the set's spam and ham, respectively; hence the
grouping. For example, EDS SAe-46 is comprised of 400 spam mails
taken from the group SAs-G2 and 600 ham mails from the group
SAe-G2, and EDS BKS-SAeh-15 is comprised of 200 spam mails from
the BKS data set and 1000 ham mails made up of 800 mails from the
SAe-G2 group and 200 mails from the SAh group.

\subsubsection{Pre-processing}
\label{SubSubSection Pre-processing}

For the suffix tree classifier, no pre-processing is done. It is
likely that some pre-processing of the data may improve the
performance of an ST classifier, but we do not address this issue
in the current paper.
\begin{flushleft}
For the the naive Bayesian classifier, we use the following standard three
pre-processing procedures:
\begin{enumerate}
\item Remove all punctuation. \item Remove all stop-words. \item
Stem all remaining words.
\end{enumerate}
Words are taken as strings of characters separated from other
strings by one or more whitespace characters (spaces, tabs,
newlines). Punctuation is removed first in the hope that many of
the intra-word characters which spammers use to confuse a Bayesian
filter will be removed. Our stop-word list consisted of the 57 of
the most frequent prepositions, pronouns, articles and
conjunctives. Stemming was done using an implementation of
Porter's 1980 algorithm, more recently reprinted
in~\cite{Porter97}. All words less than three characters long are ignored. For more general information on these and
other approaches to pre-processing, the reader is directed
to~\cite{manning99foundationsSNLP,wiess05textmining}.
\end{flushleft}

\subsection{Performance Measurement}
\label{SubSection Performance Measurement}

There are generally two sets of measures used in the literature;
here we introduce both in order that our results may be more
easily compared with previous work.

Following~\cite{sahami98bayesian},~\cite{androutsopoulos00evaluation}, and others, the first set of measurement parameters we use are \textit{recall} and \textit{precision} for both spam and ham. For spam (and similarly for ham) these measurements are defined as follows: \\

$Spam \hspace{1ex} Recall \hspace{1ex} (SR) = \frac{SS}{SS + SH}$
$,$\hspace{1ex} $Spam \hspace{1ex} Precision \hspace{1ex} (SP) =
\frac{SS}{SS + HS}$ \vspace{1ex} \\
where $XY$ means the number of items of class $X$ assigned to
class $Y$; with $S$ standing for spam and $H$ for ham. Spam recall
measures the proportion of all spam messages which were identified
as spam and spam precision measures the proportion of all messages
classified as spam which truly are spam; and similarly for ham.

However, it is now more popular to
measure performance in terms of \textit{true positive} (TP) and
\textit{false positive} (FP) rates:
\begin{center}
$TPR = \frac{SS}{SS + SH}$ $,$\hspace{4ex} $FPR = \frac{HS}{HH + HS}$ \vspace{2ex} \\
\end{center}
The TPR is then the proportion of spam correctly classified as
spam and the FPR is the proportion of ham incorrectly classified
as spam. Using these measures, we plot in Section~\ref{Section
Results} what are generally referred to as receiver operator curves (ROC)
\cite{fawcett04roc} to observe the behaviour of the classifier at
a range of thresholds.

To precisely see performance rates for particular thresholds, we
also found it useful to plot, against threshold, false positive
rates (FPR) and false negative rates (FNR):
\begin{center}
$FNR = 1- TPR$ \vspace{2ex} \\
\end{center}

Effectively, FPR measures errors in the classification of ham and
FNR measures errors in the classification of spam.

\section{Results}
\label{Section Results}

We begin in Section~\ref{SubSection Assessment} by comparing the
results of the suffix tree (ST) approach to the reported results
for a naive Bayesian (NB) classifier on the the Ling Spam corpus. We
then extend the investigation of the suffix tree to other data
sets to examine its behaviour under different conditions and
configurations. To maintain a comparative element on the further
data sets we implemented an NB classifier which proved
to be competitive with the classifier performance as reported in
\cite{androutsopoulos00evaluation} and others. In this way we look at each
experimental parameter in turn and its effect on the performance
of the classifier under various configurations.

\subsection{Assessment}
\label{SubSection Assessment}

\setlength{\captionmargin}{28pt}
\begin{table}
\begin{center}
\caption{Results of (Androutsopoulos et al., 2000) on the
Ling-Spam corpus. In the pre-processing column: 'bare' indicates
no pre-processing. The column labelled 'No. of attrib.' indicates
the number of word features which the authors retained as
indicators of class. Results are shown at the bottom of the table
from ST classification using a linear significance function and no
normalisation; for the ST classifier, we performed no
pre-processing and no feature selection.} \vspace{0.5cm}
\begin{tabular}{|c|l|c|c|c|c|}
  \hline
  & Pre-processing & No. of & $th$  & SR(\%) & SP(\%) \\
  & & attrib. & & & \\
  \cline{1-6}
NB  &(a) bare & 50 & 1.0 & 81.10 & 96.85 \\
  &(b) stop-list & 50 & 1.0 & 82.35 & 97.13 \\
  &(c) lemmatizer & 100 & 1.0 & 82.35 & 99.02 \\
  &(d) lemmatizer + stop-list & 100 & 1.0 & 82.78 & 99.49 \\
  \cline{2-6}
  &(a) bare & 200 & 0.11 & 76.94 & 99.46 \\
  &(b) stop-list & 200 & 0.11 & 76.11 & 99.47 \\
  &(c) lemmatizer & 100 & 0.11 & 77.57 & 99.45 \\
  &(d) lemmatizer + stop-list & 100 & 0.11 & 78.41 & 99.47 \\
  \cline{2-6}
  &(a) bare & 200 & 0.001 & 73.82 & 99.43 \\
  &(b) stop-list & 200 & 0.001 & 73.40 & 99.43 \\
  &(c) lemmatizer & 300 & 0.001 & 63.67 & 100.00 \\
  &(d) lemmatizer + stop-list & 300 & 0.001 & 63.05 & 100.00 \\
  \hline
ST & bare & N/A & 1.00 & 97.50 & 99.79 \\
   \cline{2-6}
   & bare & N/A & 0.98 & 96.04 & 100.00 \\
   \hline
\end{tabular}
\label{Table Results NB+ST LS-FULL}
\end{center}
\end{table}

Table~\ref{Table Results NB+ST LS-FULL} shows the results reported
in~\cite{androutsopoulos00evaluation}, from the application of
their NB classifier on the LS-FULL data set, and the results of
the ST classifier, using a linear significance function with no
normalisation, on the same data set.

As can be seen, the performance levels for precision are
comparable, but the suffix tree simultaneously achieves much
better results for recall.

\cite{androutsopoulos00evaluation} test a number of
thresholds~\footnote{(Androutsopoulos et al. 2000) do not actually
quote the threshold, but a 'cost value', which we have converted
into its threshold equivalent.} (\textit{th}) and found that their NB filter achieves a
100\% spam precision (SP) at a threshold of 0.001. We similarly tried a number of
thresholds for the ST classifier, as previously explained (see
Section~\ref{SubSection Experimental Parameters}), and found that
100\% SP was achieved at a threshold of 0.98. Achieving high SP
comes at the inevitable cost of a lower spam recall (SR), but we
found that our ST can achieve the 100\% in SP with less cost in
terms of SR, as can be seen in the table.

As stated in the table (and previously: see
Section~\ref{SubSubSection Pre-processing}), we did no
pre-processing and no feature selection for the suffix tree.
However, both of these may well improve performance, and we intend
to investigate this in future work.

\setlength{\captionmargin}{28pt}
\begin{table}
\begin{center}
\caption{Results of in-house naïve Bayes on the LS-FULL data set,
with stop-words removed and all remaining words lemmatized. The
number of attributes was unlimited, but, for the LS-FULL data set,
in practice the spam vocabulary was approximately 12,000, and the
ham vocabulary approximately 56,000, with 7,000 words appearing in
both classes.} \vspace{0.5cm}
\begin{tabular}{|c|l|c|c|c|c|}
  \hline
  & Pre-processing & No. of & $th$  & SR(\%) & SP(\%) \\
  & & attrib. & & & \\
  \cline{1-6}
NB*  & lemmatizer + stop-list & unlimited & 1.0 & 99.16 & 97.14 \\
  & lemmatizer + stop-list & unlimited & 0.94 & 89.58 & 100.00 \\
     \hline
\end{tabular}
\label{Table Results NB* LS-FULL}
\end{center}
\end{table}

As we mentioned earlier (and in Section~\ref{Section Naive Bayesian Classification}), we use our own NB classifier in our further
investigation of the performance of our ST classifier. We
therefore begin by presenting in Table~\ref{Table Results NB* LS-FULL} the
results of this classifier (NB*) on the LS-FULL data set. As the
table shows, we found our results were, at least in some cases,
better than those reported in \cite{androutsopoulos00evaluation}.
This is an interesting result which we do not have space to
investigate fully in this paper, but there are a number of
differences in our naive Bayes method which may account for this.

Firstly \cite{androutsopoulos00evaluation} uses a maximum of 300
attributes, which may not have been enough for this domain or data
set, whereas we go to the other extreme of not limiting our number
of attributes, which would normally be expected to ultimately
reduce performance, but only against an optimal number, which is
not necessarily the number used by
\cite{androutsopoulos00evaluation}. Indeed, some researchers
\cite{DBLP:journals/cj/LiJ98,Zhang04evaluation} have found NB does
not always benefit from feature limitation, while others have
found the optimal number of features to be in the thousands or
tens of thousands \cite{Schneider03spam,mccallum98comparison}.
Secondly, there may be significant differences in our
pre-processing, such as a more effective stop-word list and
removal of punctuation; and thirdly, we estimate the probability
of word features using Laplace smoothing (see
formula~\ref{P(d_i|c_j)}), which is more robust than the estimated
probability quoted by \cite{androutsopoulos00evaluation}.

There may indeed be further reasons, but it is not our intension
in this paper to analyse the NB approach to text classification,
but only to use it as a comparative aid in our investigation of
the performance of the ST approach under various conditions.
Indeed, other researchers have extensively investigated NB and for
us to conduct the same depth of investigation would require a
dedicated paper.

\setlength{\captionmargin}{100pt}
\begin{table}
\begin{center}
\caption{Precision-recall breakeven points on the LS-FULL data
set.} \vspace{0.5cm}
\begin{tabular}{|c|c|c|}
  \hline
  Classifier & Spam (\%) & Ham (\%) \\
  \hline
  NB$'$ & 96.47 & 99.34 \\
  NB* & 94.96 & 98.82 \\
  ST & 98.75 & 99.75 \\
  \hline
\end{tabular}\label{Table Breakeven Values}
\end{center}
\end{table}

Furthermore, both our NB* and ST classifiers appear to be
competitive with quoted results from other approaches using the
same data set. For example in \cite{Schneider03spam}, the author
experiments on the Ling-Spam data set with different models of NB
and different methods of feature selection, and achieves results
approximately similar to ours. \cite{Schneider03spam} quotes
``\textit{breakeven}" points, defined as the ``highest recall for
which recall equaled precision", for both spam and ham;
Table~\ref{Table Breakeven Values} shows the results achieved by
the author's best performing naive Bayes configuration (which we
label as `NB$'$') alongside our naive Bayes (NB*) and the suffix
tree (ST) using a linear significance function and no
normalisation. As can be seen, NB* achieves slightly worse results
than the NB$'$, while ST achieves slightly better results; but all
are clearly competitive. And as a final example,
in~\cite{surkov04phd} the author applies developments and
extentions of support vector machine algorithms
\cite{vapnik99learningtheory} to the Ling-Spam data set, albeit in
a different experimental context, and achieves a minimum sum of
errors of 6.42\%; which is slightly worse than the results
achieved by our NB* and ST classifiers.

Thus, let us proceed on the assumption that both our (NB* and ST)
classifiers are at least competitive enough for the task at hand:
to investigate how their performance \textit{varies} under
experimental conditions for which results are not available in the
literature.

\subsection{Analysis}
\label{SubSection Analysis}

In the following tables, we group email data sets (EDSs), as in
Table~\ref{Table Composition Of EDSs}, Section~\ref{SubSection
Data}, by their source corpora, so that each of the EDSs in one
group differ from each other only in the proportion of spam to ham
they contain.

\subsubsection{Effect of Depth Variation}
\label{SubSubSection Effect of Depth Variation}

\setlength{\captionmargin}{114pt}
\begin{table}
\begin{center}
\caption{Classification errors by depth using a constant
significance function, with no normalisation, and a threhsold of 1
on the LS-11 email data set.} \vspace{0.5cm}
\label{Table ST Depth
Variation}
\begin{tabular} {|c||c|c|}
\hline
Depth & FPR(\%) & FNR(\%) \\
\hline
2 & 58.75  & 11.75 \\
4 & 0.25  & 4.00 \\
6 & 0.50  & 2.50 \\
8 & 0.75 & 1.50 \\
\hline
\end{tabular}
\end{center}
\end{table}

For illustrative purposes, Table~\ref{Table ST Depth Variation}
shows the results using the $constant$ significance function, with
no normalisation using the LS-11 data set. Depths of 2, 4, 6, and
8 are shown.

The table demonstrates a characteristic which is common to all
considered combinations of significance and normalisation
functions: performance improves as the depth increases. Therefore,
in further examples, we consider only our maximum depth of 8.
Notice also the decreasing marginal improvement as depth
increases, which suggests that there may exist a maximal
performance level, which was not necessarily achieved by our trials.

\subsubsection{Effect of Significance Function}
\label{SubSubSection Effect of Significance Function}
\setlength{\captionmargin}{40pt}
\begin{figure}
\begin{center}
\includegraphics[width=10cm]{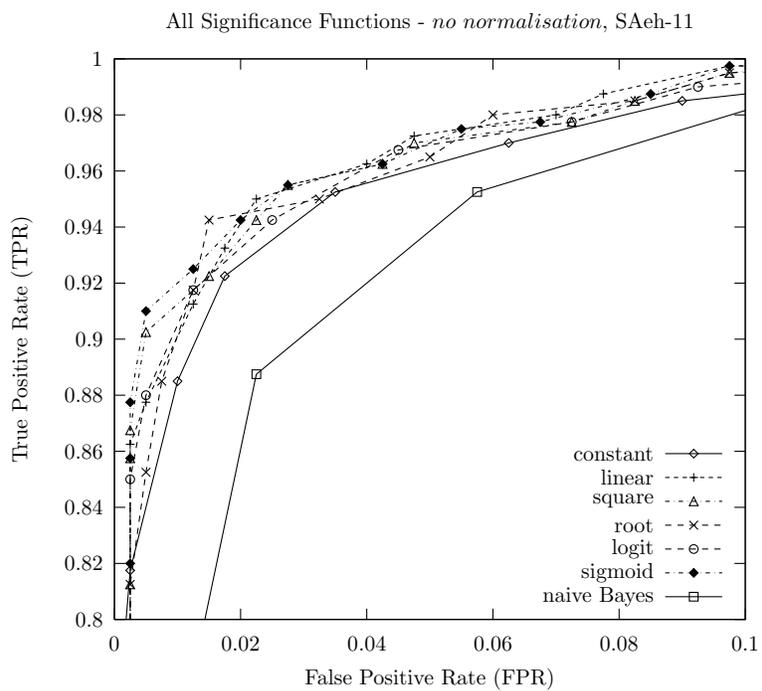}
\caption{ROC curves for all significance functions on the SAeh-11
data set.} \label{Figure ROC; All Functions; NN; SAeh-11}
\end{center}
\end{figure}

We found that all the significance functions we tested worked very
well, and all of them performed better than our naive Bayes.
Figure~\ref{Figure ROC; All Functions; NN; SAeh-11} shows the ROC
curves produced by each significance function (with no
normalisation) for what proved to be one of the most difficult
data sets (SAeh-11: see Section~\ref{SubSection Data}).

\setlength{\captionmargin}{30pt}
\begin{table}
\begin{center}
\caption{Sum of errors (FPR+FNR)values at a conventional threshold
of 1 for all significance functions under match permutation
normalisation. The best scores for each email data set are
highlighted in bold.} \vspace{0.5cm} \label{Table_Significance
Function Conventional Threshold}
\begin{tabular}{|c||c|c|c|c|c|c|c|}
\hline
 & \multicolumn{6}{|c|}{Sum of Errors (\%) at $th = 1$}  \\
 & \multicolumn{6}{|c|}{for specifications of $\phi[\hat{p}]$} \\
EDS Code & constant & linear & square & root & logit & sigmoid \\
\hline
 & & & & & & \\
LS-11 & \textbf{1.5} & 2.25 & 2.5 & 1.75 & 1.75 & 1.75 \\
LS-46 & 1.33 & 1.42 & 1.92 & \textbf{1.08} & 1.58 & 1.42 \\
LS-15 & \textbf{1.33} & \textbf{1.33} & 1.55 & \textbf{1.33} & 1.55 & 1.89 \\
 & & & & & & \\
SAe-11 & \textbf{0.25} & 0.5 & 0.5 & 0.5 & \textbf{0.25} & 0.75 \\
SAe-46 & 0.5 & 0.75 & 0.5 & 0.5 & \textbf{0.25} & 0.75 \\
SAe-15 & \textbf{1.00} & 1.50 & 1.8 & 1.5 & 1.1 & 2.00 \\
 & & & & & & \\
SAeh-11 & 7.00 & 7.00 & 7.50 & 6.75 & \textbf{5.49} & 6.50 \\
SAeh-46 & \textbf{4.33} & 4.58 & 4.92 & 5.00 & 4.42 & 4.92 \\
SAeh-15 & 9.3 & \textbf{7.5} & 8.00 & 7.7 & 7.6 & 8.6 \\
 & & & & & & \\
BKS-LS-11 & \textbf{0} & \textbf{0} & \textbf{0} & \textbf{0} & \textbf{0} & \textbf{0} \\
BKS-LS-46 & \textbf{0} & \textbf{0} & \textbf{0} & \textbf{0} & \textbf{0} & \textbf{0} \\
BKS-LS-15 & \textbf{0} & 1.5 & 1.5 & 1.00 & \textbf{0} & 1.5 \\
 & & & & & & \\
BKS-SAe-11 & 4.75 & 1.75 & \textbf{1.5} & \textbf{1.5} & \textbf{1.5} & 1.75 \\
BKS-SAe-46 & 4.5 & 1.75 & 2.00 & 1.75 & \textbf{1.5} & 2.75 \\
BKS-SAe-15 & 9.5 & 6.00 & 6.00 & \textbf{5.50} & \textbf{5.50} & 8.5 \\
 & & & & & & \\
BKS-SAeh-11 & 9.25 & 5.75 & 7.25 & \textbf{5.00} & 5.75 & 7.25 \\
BKS-SAeh-46 & 10.25 & 5.25 & 7.00 & \textbf{4.25} & 5.00 & 7.25 \\
BKS-SAeh-15 & 15.5 & \textbf{9.5} & \textbf{9.5} & \textbf{9.5} & \textbf{9.5} & 14.5 \\
 & & & & & & \\
\hline
\end{tabular} \\
\end{center}
\end{table}
%\vspace{0.8ex}

\setlength{\captionmargin}{30pt}
\begin{table}
\begin{center}
\caption{Sum of error (FPR+FNR) values at individual optimal
thresholds for all significance functions under match permutation
normalisation. The best scores for each data set are highlighted
in bold.}\vspace{0.5cm}\label{Table Significance Function Optimal
Threshold}
\begin{tabular} {|c||c|c|c|c|c|c|c|}
\hline
 & \multicolumn{6}{|c|}{Sum of Errors (\%) at optimal $th$}\\
 & \multicolumn{6}{|c|}{for specifications of $\phi[\hat{p}]$}  \\
EDS Code & constant & linear & square & root & logit & sigmoid \\
\hline
 & & & & & & \\
LS-11 & 1.25 & \textbf{1.00} & \textbf{1.00} & \textbf{1.00} & \textbf{1.00} & \textbf{1.00}\\
LS-46 & 1.08 & 1.08 & 1.00 & 1.08 & 1.08 & \textbf{0.83}\\
LS-15 & \textbf{1.33} & \textbf{1.33} & \textbf{1.33} & \textbf{1.33} & \textbf{1.33} & \textbf{1.33} \\
 & & & & & &\\
SAe-11 & 0.25 & \textbf{0} & \textbf{0} & \textbf{0} & \textbf{0} & 0.25\\
SAe-46 & 0.42 & 0.33 & 0.33 & \textbf{0.25} & \textbf{0.25} & 0.5\\
SAe-15 & \textbf{1.00} & 1.3 & 1.4 & 1.1 & 1.1 & 1.2\\
 & & & & & &\\
SAeh-11 & 7.00 & 6.50 & \textbf{6.00} & 6.25 & 6.25 & 6.50\\
SAeh-46 & \textbf{4.00} & 4.58 & 4.92 & 4.42 & 4.33 & 4.92\\
SAeh-15 & 6.50 & 6.60 & 6.70 & 6.50 & 6.60 & \textbf{6.30}\\
 & & & & & & \\
BKS-LS-11 & \textbf{0} & \textbf{0} & \textbf{0} & \textbf{0} & \textbf{0} & \textbf{0}\\
BKS-LS-46 & \textbf{0} & \textbf{0} & \textbf{0} & \textbf{0} & \textbf{0} & \textbf{0}\\
BKS-LS-15 & \textbf{0} & \textbf{0} & \textbf{0} & \textbf{0} & \textbf{0} & \textbf{0}\\
 & & & & & & \\
BKS-SAe-11 & \textbf{0} & \textbf{0} & \textbf{0} & \textbf{0} & \textbf{0} & \textbf{0}\\
BKS-SAe-46 & \textbf{0} & \textbf{0} & \textbf{0} & \textbf{0} & \textbf{0} & \textbf{0}\\
BKS-SAe-15 & 0.2 & \textbf{0} & \textbf{0} & \textbf{0} & 0.2 & 0.50\\
 & & & & & & \\
BKS-SAeh-11 & 2.75 & \textbf{1.75} & 2.00 & \textbf{1.75} & 2.00 & 2.00\\
BKS-SAeh-46 & 1.33 & 1.17 & 1.50 & 1.17 & \textbf{1.00} & 1.33\\
BKS-SAeh-15 & \textbf{1.1} & 1.2 & 2.00 & 1.30 & \textbf{1.1} & 2.1\\
 & & & & & &\\
\hline
\end{tabular} \\
\end{center}
\end{table}

We found little difference between the performance of each of the
functions across all the data sets we experimented with, as can be
seen from the summary results in Table~\ref{Table_Significance
Function Conventional Threshold}, which shows the minimum sum of
errors (FPR+FNR) achieved at a threshold of 1.0 by each significance
function on each data set. The constant function looks marginally
the worst performer and the logit and root functions marginally
the best, but this difference is partly due to differences in
optimal threshold (see Section~\ref{SubSubSection Effect of
Threshold Variation}) for each function: those that perform less
well at a threshold of 1.0 may perform better at other thresholds.

Table~\ref{Table Significance Function Optimal Threshold} presents
the minimum sum of errors achieved by each function at its
individual optimal threshold. In this table there is even less
difference between the functions, but still the root looks
marginally better than the others, in that it appears to most frequently achieve the lowest sum of errors, and so, for the sake of brevity
we favour this function in much of our following analysis.

\subsubsection{Effect of Threshold Variation}
\label{SubSubSection Effect of Threshold Variation} We generally
found that there was an optimal threshold (or range of thresholds)
which maximised the success of the classifier. As can be seen from
the four example graphs shown in Figure~\ref{Figure Probability
Function Examples}, the optimal threshold varies depending on the
significance function and the mix of ham and spam in the training
and testing sets, but it tends to always be close to 1.

Obviously, it may not be possible to know the optimal threshold in
advance, but we expect, though have not shown, that the optimal
threshold can be established during a secondary stage of training
where only examples with scores close to the threshold are used -
similar to what~\cite{meyer04spambayes} call ``non-edge training".

In any case, the main reason for using a threshold is to allow a
potential user to decide the level of false positive risk they are
willing to take: reducing the risk carries with it an inevitable
rise in false negatives. Thus we may consider the lowering of the
threshold as attributing a greater cost to miss-classified ham
(false positives) than to miss-classified spam; a threshold of 1.0
attributes equal importance to the the two.

The shapes of the graphs are typical for all values of
$\phi[\hat{p}]$; the performance of a particular scoring
configuration is reflected not only by the minimums achieved at
optimal thresholds but also by the steepness (or shallowness) of
the curves: the steeper they are, the more rapidly errors rise at
sub-optimal levels, making it harder to achieve zero false
positives without a considerable rise in false negatives. Graph
(d) shows that our NB classifier is the most unstable in this
respect. \setlength{\captionmargin}{40pt}
\begin{figure}
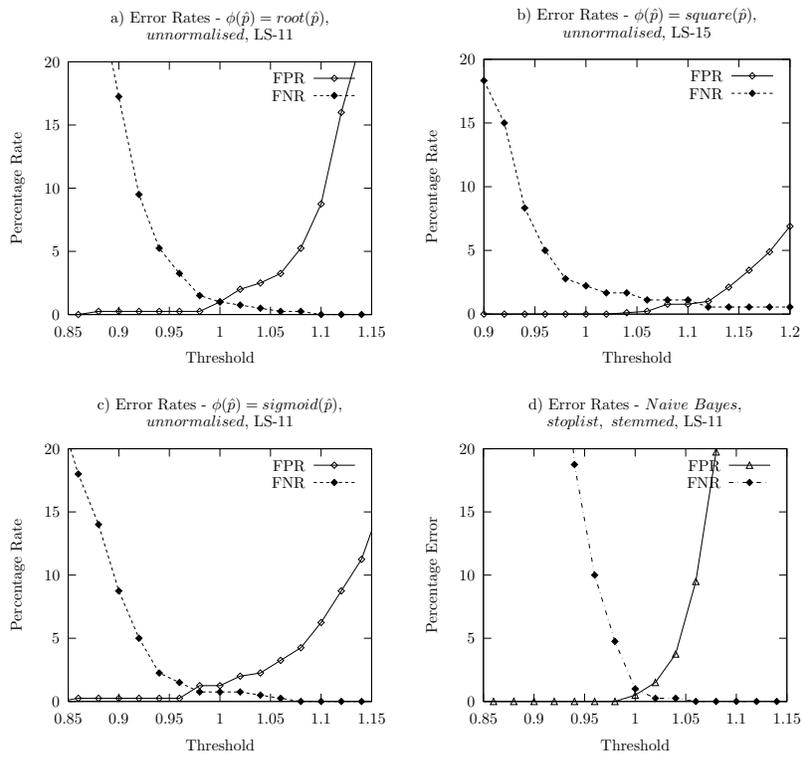

\begin{center}
\begin{tabular}{ll}
\includegraphics[width=5cm]{graphics1/th_FPFN_LS-11_8300.8}
\vspace{0.3cm} &
\includegraphics[width=5cm]{graphics1/th_FPFN_LS-15_8200.11} \\
\includegraphics[width=5cm]{graphics1/th_FPFN_LS-11_8500.14}
\vspace{0.3cm} &
\includegraphics[width=5cm]{graphics1/th_FPFN_NB_LS-11_Stop+Lemm.17} \\
\end{tabular}
\caption{Effect of threshold variation. Graphs (a-c) show suffix tree false positive (FP) and false negative (FN) rates for three specification of $\phi(\hat{p})$ under no normalisation; graph (d) shows naive Bayes FP and FN rates.} \label{Figure Probability Function
Examples}
\end{center}
\end{figure}

\subsubsection{Effect of Normalisation}
\label{SubSubSection Effect of Normalisation} 
\begin{figure}
\begin{center}
\includegraphics[width=10cm]{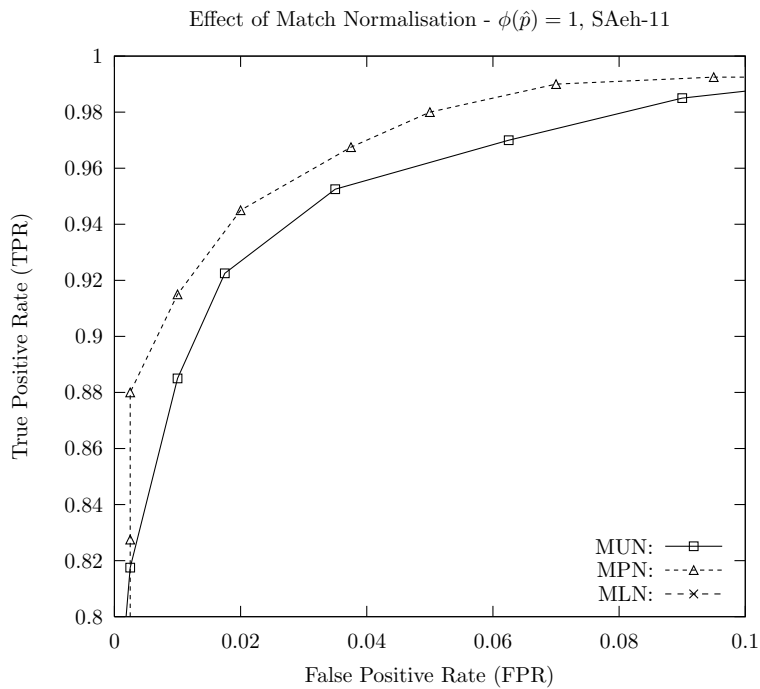}
\caption{ROC curves for the constant significance function under
no match normalisation (MUN), match permutation normalisation
(MPN) and match length normalisation (MLN), with no tree
normalisation, on the SAeh-11 data set. MLN has such a detrimental
effect on performance that its ROC curve is off the scale of the
graph.} \label{Figure ROC phi(p)=1; All MN; SAeh-11}
\end{center}
\end{figure}

We found that there
was a consistent advantage to using \textit{match permutation}
normalisation, which was able to improve overall performance as
well as making the ST classifier more stable under varying
thresholds. Figure~\ref{Figure ROC phi(p)=1; All MN; SAeh-11}
shows the ROC curves produced by the constant significance
function under match permutation normalisation (MPN); match length
normalisation (MLN) reduced performance so much that the resulting
curve does not even appear in the range of the graph. The
stabilising effect of match permutation normalisation is reflected
in ROC curves by an increase in the number of points along the
curve, but may be better seen in Figure~\ref{Figure Effect Of
Match Normalisation} as a shallowing of the FPR and FNR curves.
The negative effect of MLN concurs with our heuristics from
Section~\ref{SubSection Classification using Suffix Trees}.
\setlength{\captionmargin}{40pt}
\setlength{\captionmargin}{40pt}
\begin{figure}
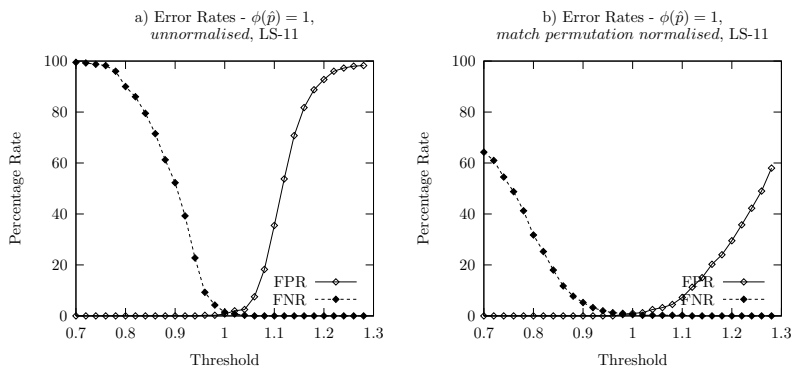

\begin{center}
\begin{tabular}{ll}
\includegraphics[width=5cm]{graphics1/th_FPFN_LS-11_8000.22} &
\includegraphics[width=5cm]{graphics1/th_FPFN_LS-11_8010.24}
\end{tabular}
\caption{Effect of match permutation normalisation. False positive (FP)
and false negative (FN) rates using a constant significance function on the LS-11 EDS. Graph (a) shows the false positive (FP) and false negative (FN) rates under no normalisation and graph (b) shows FP and FN rates under match permutation normalisation.} \label{Figure Effect Of Match Normalisation}
\end{center}
\end{figure}

\subsubsection{Effect of Spam to Ham Ratios}
\label{SubSubSection Effect of Spam to Ham Ratios}

We initially found that the mix of spam to ham in the data sets
could have some effect on performance, with the degree of
difference in performance depending on the data set and the
significance function used; however, with further investigation we
found that much of the variation was due to differences in the
optimal threshold. This can be seen by first examining the
differences in performance for different spam:ham ratios shown in
Table~\ref{Table_Significance Function Conventional Threshold}, in
which a 1:5 ratio appears to result in lower performance than the
more balanced ratios of 4:6 and 1:1; then examining the results
presented in Table~\ref{Table Significance Function Optimal
Threshold}, where differences are far less apparent. These
observations are reinforced by the graphs shown in
Figure~\ref{Figure Differences In Ratios SAeh}. In graph (a) which
shows the ROC curves produced by the constant significance
function with no normalisation on the SAeh data sets, we can see
that the curves produced by different ratios appear to achieve
slightly different maximal performance levels but roughly follow
the the same pattern. Graphs (b-c) further show that the maximal
levels of performance are achieved at different threshold for each
ratio.

\setlength{\captionmargin}{40pt}
\begin{figure}
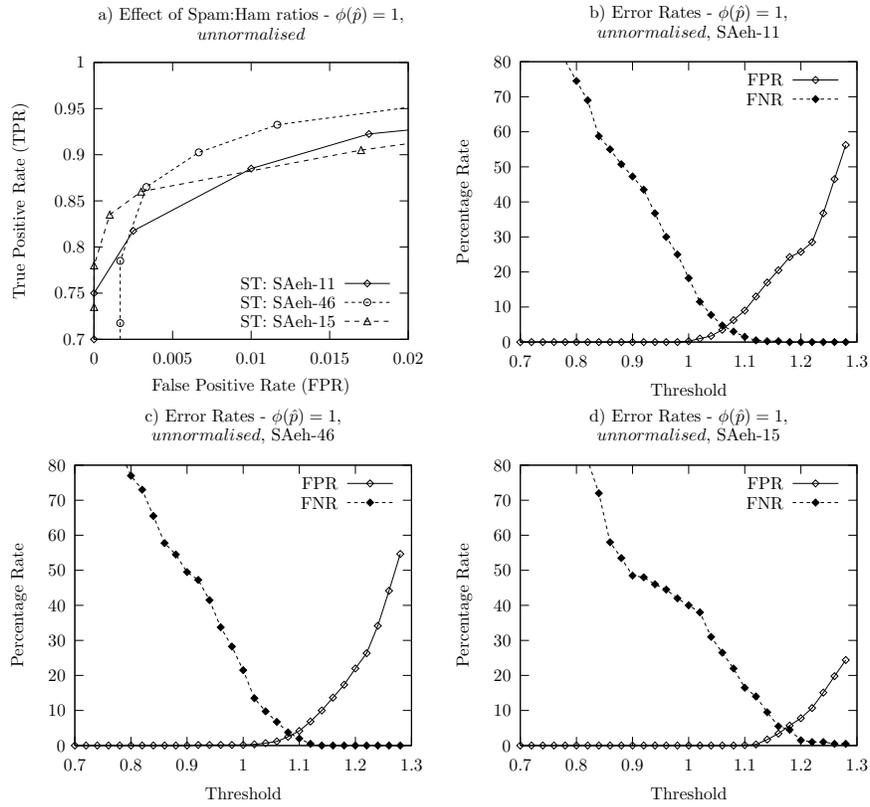

\begin{center}

\begin{tabular}{ll}
\includegraphics[width=5.5cm]{graphics1/ROC8000_SAeh-11,46,15.35} &
\includegraphics[width=5.5cm]{graphics1/th_FPFN_SAeh-11_8000.34} \\
\includegraphics[width=5.5cm]{graphics1/th_FPFN_SAeh-46_8000.33} &
\includegraphics[width=5.5cm]{graphics1/th_FPFN_SAeh-15_8000.32} \\
\end{tabular}
\caption{Effect of varying ratios of spam:ham on the SAeh data
using a constant significance function with no normalisation.
Graph (a) shows the ROC curves produced for each ratio; while
graphs (b-d) show the FP and FN rates separately for ratios of
1:1, 4:6 and 1:5 respectively.} \label{Figure Differences In
Ratios SAeh}
\end{center}
\end{figure}

\subsubsection{Overall Performance Across Email Data Sets}
\label{SubSubSection Overall Performance Across Email Data Sets}
\setlength{\captionmargin}{40pt}
\begin{table}
\begin{center}
\caption{Classification errors at threshold of 1, for Naive Bayes
(NB) and a Suffix Tree (ST) using a root significance function and
match permutation normalisation, but no tree
normalisation.}\vspace{0.5cm} \label{Table ST vs NB All EDSs}
\begin{tabular} {|c||c|c|c|c|}
\hline
 & \multicolumn{2}{|c|}{Naive Bayes} &
 \multicolumn{2}{|c|}{Suffix Tree} \\
EDS Code & FPR (\%) & FNR (\%) & FPR (\%) & FNR (\%) \\
\hline
 & & & &  \\
LS-11 & 1.25 & 0.50 & 1.00 & 1.75 \\
LS-46 & 0.67 & 1.25 & 0.83 & 0.25 \\
LS-15 & 1.00 & 1.00 & 0.22 & 0.13  \\
 & & & &  \\
SAe-11 & 0 & 2.75 & 0 & 0.50 \\
SAe-46 & 0.17 & 2.00 & 0 & 0.50 \\
SAe-15 & 0.30 & 3.50 & 0 & 1.50 \\
 & & & &  \\
SAeh-11 & 10.50 & 1.50 & 3.50 & 3.25 \\
SAeh-46 & 5.67 & 2.00 & 2.00 & 3.00 \\
SAeh-15 & 4.10 & 7.00 & 0.70 & 7.00 \\
 & & & &  \\
BKS-LS-11 & 0 & 12.25 & 0 & 0 \\
BKS-LS-46 & 0.17 & 13.75 & 0 & 0 \\
BKS-LS-15 & 0.20 & 30.00 & 0 & 1.00 \\
 & & & &  \\
BKS-SAe-11 & 0 & 9.00 & 0 & 1.50 \\
BKS-SAe-46 & 0 & 8.25 & 0 & 1.75 \\
BKS-SAe-15 & 1.00 & 15.00 & 0 & 5.5 \\
 & & & & \\
BKS-SAeh-11 & 16.50 & 0.50 & 0 & 5.00 \\
BKS-SAeh-46 & 8.17 & 0.50 & 0 & 4.25 \\
BKS-SAeh-15 & 8.10 & 5.50 & 0 & 9.50 \\
 & & & &  \\
\hline
\end{tabular}
\end{center}
\end{table}

\setlength{\captionmargin}{0pt}
\begin{table}
\begin{center}
\caption{Classification Errors at optimal thresholds (where the
sum of the errors is minimised) for Naive Bayes (NB) and a Suffix
Tree (ST) using a root significance function and match permutation
normalisation, but no tree normalisation.}\vspace{0.5cm}
\label{Table ST vs NB Optimal Threshold All EDSs}
\begin{tabular} {|c||c|c|c|c|c|c|}
\hline
 & \multicolumn{3}{|c|}{Naive Bayes} &
 \multicolumn{3}{|c|}{Suffix Tree} \\
EDS Code & OtpTh & FPR (\%) & FNR (\%) & OptTh & FPR (\%) & FNR (\%) \\
\hline
 & & & & & & \\
LS-11 & 1.0 & 1.25 & 0.50 & 0.96 & 0 & 1.00 \\
LS-46 & 1.02 & 1.00 & 0.67 & 0.96 & 0.33 & 0.75 \\
LS-15 & 1.00 & 1.00 & 1.00 & 0.98 - 1.00 & 0.22 & 1.11 \\
 & & & & & & \\
SAe-11 & 1.06 & 0.25 & 0 & 1.10 & 0 & 0 \\
SAe-46 & 1.04 & 0.33 & 0.25 & 1.02 & 0 & 0.25 \\
SAe-15 & 1.02 & 2.30 & 1.50 & 1.02 & 0.1 & 1.00 \\
 & & & & & & \\
SAeh-11 & 0.98 & 10.50 & 1.50 & 0.98 & 2.75 & 3.50 \\
SAeh-46 & 1.00 & 5.67 & 2.00 & 0.98 & 1.16 & 3.25 \\
SAeh-15 & 1.02 & 7.60 & 1.50 & 1.10 & 3.50 & 3.00 \\
 & & & & & & \\
BKS-LS-11 & 1.04 & 0.75 & 2.25 & 0.78 - 1.22 & 0 & 0 \\
BKS-LS-46 & 1.06 & 2.50 & 1.25 & 0.78 - 1.16 & 0 & 0 \\
BKS-LS-15 & 1.10 & 5.50 & 1.50 & 1.02 - 1.22 & 0 & 0 \\
 & & & & & & \\
BKS-SAe-11 & 1.04 - 1.06 & 0 & 0.25 & 1.04 - 1.28 & 0 & 0 \\
BKS-SAe-46 & 1.06 & 0.50 & 0.25 & 1.18 - 1.28 & 0 & 0 \\
BKS-SAe-15 & 1.04 & 6.90 & 0 & 1.20 & 0 & 0 \\
 & & & & & &\\
BKS-SAeh-11 & 0.98 & 8.00 & 2.00 & 1.06 & 0 & 1.75 \\
BKS-SAeh-46 & 0.98 & 4.00 & 3.75 & 1.14 - 1.16 & 0.67 & 0.5 \\
BKS-SAeh-15 & 1.00 & 8.10 & 5.50 & 1.24 - 1.26 & 0.80 & 0.50 \\
 & & & & & & \\
\hline
\end{tabular}
\end{center}
\end{table}

Table~\ref{Table ST vs NB All EDSs} summarises the results for
both the ST and NB classifiers at a threshold of 1.0 and
Table~\ref{Table ST vs NB Optimal Threshold All EDSs} summarises
results at the individual optimal thresholds which minimise the
sum of the errors (FPR+FNR).

We found that the performance of the NB is in some cases dramatically improved at its optimal threshold, for example in the case of the LS-BKS data sets. But at both a threshold of 1.0 and at optimal thresholds, the NB classifier behaves very much as expected,
supporting our initial assumptions as to the difficulty of the
data sets. This can be clearly seen in Table~\ref{Table ST vs NB Optimal Threshold All EDSs}: on the SAeh data sets which contain ham with 'spammy'
features, the NB classifier's false positive rate increases,
meaning that a greater proportion of ham has been incorrectly
classified as spam; and on the BKS-SAeh data sets which
additionally contain spam which is disguised to appear as ham,
the NB classifier's false negative rate increases, meaning that a
greater proportion of spam has been misclassified as ham.

The performance of the ST classifier also improves at its optimal thresholds, though not so dramatically, which is to be expected considering our understanding of how it response to changes in the threshold (see Section~\ref{SubSubSection Effect of Threshold Variation}). The ST also shows improved performance on data sets involving BKS data. This may be because the character level analysis of the suffix tree approach is able to treat the attempted obfuscations as further positive distinguishing features, which do not exist in the more standard examples of spam which constitute the LS data sets. In all cases except on the SAeh data, the ST is able to keep the sum of errors close to or below 1.0, and in some cases, it is able to achieve a zero sum of errors. Furthermore, the suffix tree's optimal performance is often achieved at a range of thresholds, supporting our earlier observation of greater stability in it's classification success.

\subsubsection{Computational Performance}
\label{SubSubSection Computational Performance}

\setlength{\captionmargin}{20pt}
\begin{table}
\begin{center}
\caption{Computational performance of suffix tree classification
on four bare (no pre-processing) data sets. Experiments were run
on a pentium IV 3GHz Windows XP laptop with 1GB of RAM. Averages
are taken over all ten folds of cross-validation.}\vspace{0.5cm}
\label{Table Computational Performance ST}
\begin{tabular} {|c||c|c|c|c|c|}
\hline
EDS Code (size) & Training & AvSpam & AvHam & AvPeakMem \\
\hline
 & & & & \\
LS-FULL (7.40MB) & 63s & 843ms & 659ms & 765MB \\
 & & & & \\
LS-11 (1.48MB) & 36s & 221ms & 206ms & 259MB \\
 & & & & \\
SAeh-11 (5.16MB) & 155s & 504ms & 2528ms & 544MB \\
 & & & & \\
BKS-LS-11 (1.12MB) & 41s & 161ms & 222ms & 345MB \\
 & & & & \\
\hline
\end{tabular}
\end{center}
\end{table}

For illustrative purposes, in this section we provide some
indication of the time and space requirements of the suffix tree (ST)
classifier using a suffix tree of depth, $d=8$. However, it should be stressed that in our
implementation of the ST classifiers we made no attempts to
optimise our algorithms as performance was not one of our concerns
in this paper. The figures quoted here may therefore be taken as
indicators of worst-case performance levels.

Table~\ref{Table Computational Performance ST} summarises the time
and space requirements of the suffix tree classifier on four of
our email data sets. The suffix tree approach clearly and
unsurprisingly has high resource demands, far above the demands of
a naive Bayes classifier which on the same machine typically uses
no more than 40MB of memory and takes approximately 10 milliseconds (ms)
to make a classification decision.

The difference in performance across the data sets is, however,
exactly as we would expect considering our assumptions regarding
them. The first point to note is that the mapping from data set
size to tree size is non-linear. For example, the LS-FULL EDS is 5
times larger than the LS-11 EDS but results in a tree only 2.95
times larger. This illustrates the logarithmic growth of the tree
as more information is added: the tree only grows to reflect the
diversity (or complexity) of the training data it encounters and
not the actual size of the data. Hence, though the BKS-LS-11 EDS
is in fact approximately 25\% smaller than the LS-11 data set, it
results in a tree that is over 30\% larger. We would therefore
expect to eventually reach a stable maximal size once most of the
complexity of the profiled class is encoded.

The current space and time requirements are viable, though
demanding, in the context of modern computing power, but a
practical implementation would obvious benefit from optimisation
of the algorithms~\footnote{The literature on suffix trees deals
extensively with improving (reducing) the resource demands of
suffix trees \cite{ukkonen97constructing,giegerich97from,kurtz99reducing}.}.

Time could certainly be reduced very simply by implementing, for
example, a binary search over the children of each node; the
search is currently done linearly over an alphabet of
approximately 170 characters (upper- and lower- case characters
are distinguished, and all numerals and special characters are
considered; the exact size of the alphabet depends on the specific
content of the training set). And there are several other
similarly simple optimisations which could be implemented.

However, even with a fully optimised algorithm, the usual
trade-off between resources and performance will apply. With
regard to this, an important observation is that resource demands
increase exponentially with depth, whereas performance increases
logarithmically. Hence an important factor in any practical
implementation will be the choice of the depth of the suffix tree
profiles of classes.

\section{Conclusion}
\label{Section Conclusion}

Clearly, the non-parametric suffix tree performs universally well
across all the data sets we experimented with, but there is still
room for improvement: whereas in some cases, the approach is able
to achieve perfect classification accuracy, this is not
consistently maintained. Performance may be improved by
introducing some pre-processing of the data or post-processing of
the suffix tree profile, and we intend to investigate this in
future work. Certainly, the results presented in this paper
demonstrate that the ST classifier is a viable tool in the domain
of email filtering and further suggests that it may be useful in
other domains. However, this paper constitutes an initial
exploration of the approach and further development and testing is
needed.

In the context of the current work, we conclude that the choice of
significance function is the least important factor in the success
of the ST approach because all of them performed acceptably well. Different
functions will perform better on different data sets, but the root
function appeared to perform marginally more consistently well on all the
email data sets we experimented with.

Match permutation normalisation was found to be the most effective
method of normalisation and was able to improve the performance of
all significance functions. In particular it was able to improve
the success of the filter at all threshold values. However, other
methods of normalisation were not always so effective, with some
of them making things drastically worse.

The threshold was found to be a very important factor in the
success of the filter. So much so, that the differences in the
performances of particular configurations of the filter were often
 attributable more to differences in their corresponding optimal
thresholds than to the configurations themselves. However, as a
cautionary note, variations in the optimal threshold may be due to
peculiarities of the data sets involved, and this could be
investigated further.

In the case of both the NB and ST filters, it is clear that
discovering the optimal threshold -- if it were possible -- is a
good way of improving performance. It may be possible to do this
during an additional training phase in which we use some
proportion of the training examples to test the filter and
adjust the threshold up or down depending on the outcome of each
test. Of course, the threshold may be continuously changing, but
this could be handled to some extent dynamically during the actual
use of the filter by continually adjusting it in the light of any
mistakes made. This would certainly be another possible line of
investigation.

We also found that the false positive rate (FPR) and false
negative rate (FNR) curves created by varying the threshold, were
in all cases relatively shallower for our ST classifier than those
for our NB classifier, indicating that the former always performs
relatively better at non-optimal thresholds, thereby making it
easier to minimise one error without a significant cost in terms
of the other error.

Finally, any advantages in terms of accuracy in using the suffix
tree to filter emails, must be balanced against higher
computational demands. In this paper, we have given little
attention to minimising this factor, but even though available
computational power tends to increases dramatically, cost will
nevertheless be important when considering the development of the
method into a viable email filtering application, and this is
clearly a viable line of further investigation. However, the
computational demands of the approach are not intractable, and a
suffix tree classifier may be valuable in situations where
accuracy is the primary concern.

\nocite{sahami98bayesian} \nocite{meyer04spambayes} \nocite{wiess05textmining}

\vspace{5ex}

\noindent \textbf{Acknowledgments} \\

\noindent The authors are grateful to anonymous referees for
multiple comments and suggestions. The authors also wish to thank
the Police Information Technology Organisation (PITO) for its
support.

\end{document}